\documentclass{article}

\usepackage{arxiv}
\usepackage{titlesec}
\usepackage[utf8]{inputenc} 
\usepackage[T1]{fontenc}    
\usepackage{hyperref}       
\usepackage{url}            
\usepackage{booktabs}       
\usepackage{amsfonts}       
\usepackage{nicefrac}       
\usepackage{microtype}      
\usepackage{multirow}
\usepackage{graphicx}
\usepackage[numbers]{natbib}
\usepackage{doi}
\usepackage{amsmath}
\usepackage{algorithm}
\usepackage{algorithmic}
\usepackage{adjustbox}
\usepackage{cases}

\usepackage{diagbox}
\usepackage{stackengine}

\title{Quantum-Enhanced Forecasting: Leveraging Quantum Gramian Angular Field and CNNs for Stock Return Predictions}

\author{
    Zhengmeng Xu\textsuperscript{a}, Yujie Wang\textsuperscript{a}, Xiaotong Feng\textsuperscript{a}, Yilin Wang\textsuperscript{a}, Yanli Li\textsuperscript{b}, Hai Lin\textsuperscript{c,*} \\
    \textsuperscript{a}Postdoctoral Research Center, Industrial and Commercial Bank of China, Beijing, 100140, China \\
    \textsuperscript{b}School of Electrical and Information Engineering, The University of Sydney, Sydney, 2008, Australia \\
    \textsuperscript{c}School of Big Data Science, Hebei Finance University, Baoding, 071051, Hebei, China
}

\hypersetup{
pdftitle={Quantum-Enhanced Forecasting: Leveraging Quantum Gramian Angular Field and CNNs for Stock Return Predictions},
pdfsubject={q-fin.CP},
pdfauthor={Zhengmeng Xu, Hai Lin},
pdfkeywords={Time series forecasting, Quantum Gramian angular field, Convolutional neural network, Stock return predictions},
}

\begin{document}
\maketitle

\begin{abstract}
We propose a time series forecasting method named Quantum Gramian Angular Field (QGAF). This approach merges the advantages of quantum computing technology with deep learning, aiming to enhance the precision of time series classification and forecasting. We successfully transformed stock return time series data into two-dimensional images suitable for Convolutional Neural Network (CNN) training by designing specific quantum circuits. Distinct from the classical Gramian Angular Field (GAF) approach, QGAF's uniqueness lies in eliminating the need for data normalization and inverse cosine calculations, simplifying the transformation process from time series data to two-dimensional images. To validate the effectiveness of this method, we conducted experiments on datasets from three major stock markets: the China A-share market, the Hong Kong stock market, and the US stock market. Experimental results revealed that compared to the classical GAF method, the QGAF approach significantly improved time series prediction accuracy, reducing prediction errors by an average of 25\%  for Mean Absolute Error (MAE) and 48\%  for Mean Squared Error (MSE). This research confirms the potential and promising prospects of integrating quantum computing with deep learning techniques in financial time series forecasting.
\end{abstract}


\section{Introduction}
Time series analysis is a central topic in data science, primarily because of the ubiquity of time series data in diverse areas such as human activities\cite{yang2015deep}, environmental processes\cite{crozier2011using}, and biological systems\cite{bar2012studying}. This data captures how specific variables change, offering insights into various phenomena. A deeper understanding of time series can reveal trends, relationships, and patterns. Among its many applications, financial sectors heavily rely on time series data, especially for predicting stock returns, which plays a crucial role in economic planning and investment decision-making\cite{kim2003financial,wang2015forecasting}.

The task of classifying time series data presents unique challenges. Its sequential nature often makes it difficult for standard machine learning algorithms, which are generally designed for non-sequential, static data\cite{ay2022study}. This has pushed for new algorithms that can effectively handle the temporal aspects of time series data.

Recent advancements in deep learning, especially in areas like computer vision (CV)\cite{chai2021deep} and natural language processing (NLP)\cite{otter2020survey}, have provided valuable techniques for time series analysis\cite{fawaz2019deep}. Deep learning's strength lies in its ability to learn complex feature representations from data, making it suitable for handling the intricacies of time series classification. Current methods often use the raw sequence data directly or convert it into image-like formats, subsequently using tools like CNNs for processing\cite{alzubaidi2021review}.

One such image conversion technique is the Gramian Angular Field (GAF)\cite{wang2015imaging}. By turning one-dimensional time series data into two-dimensional images, GAF allows the application of image processing techniques to time series data\cite{hsu2021using}. While it can effectively capture relationships between different points in time, GAF has limitations, including the risk of losing some information and requiring significant computational resources.

On another front, quantum computing is emerging as a powerful tool that can revolutionize many areas, including machine learning\cite{biamonte2017quantum,schuld2015introduction,cerezo2022challenges}. The unique properties of quantum bits, or qubits, such as superposition and entanglement, allow for enhanced computational capabilities\cite{bennett2000quantum}. This has spurred interest in investigating the potential benefits of quantum computing for time series representation.

In this context, our study presents the QGAF, a new method for visualizing time series data. Combined with a CNN, this approach provides a comprehensive solution for predicting stock market movements. Our research introduces QGAF as a novel visualization technique, integrates it within a deep learning framework, and provides empirical evidence to support its efficacy, especially compared to the traditional GAF approach.

\section{Related Work}
\label{sec:headings}

\subsection{Machine Learning Models for Financial Time Series
Forecasting}
Financial time series data exhibit nonlinearity, dynamics, and chaos characteristics, which pose challenges for forecasting\cite{kara2011predicting}. In recent years, machine learning methods have been widely applied and proven successful in financial time series prediction, being recognized as the most effective predictive model currently available\cite{tang2022survey}. Commonly used machine learning methods include artificial neural networks, support vector machines, and others\cite{kara2011predicting,tay2001application,cao2001financial}. Artificial neural networks learn the dependencies between inputs and outputs through error backpropagation. They can handle nonlinear relationships among multiple variables, making them suitable for modeling the complex dynamics of financial markets \cite{wang2015forecasting}. Support vector machines, on the other hand, use kernel functions to map original samples into a high-dimensional feature space for linear separation, effectively dealing with the nonlinearity in financial time series data\cite{huang2005forecasting}. Meanwhile, deep learning models such as LSTM and CNN have demonstrated strong modeling capabilities in predicting market fluctuations and stock prices due to their ability to extract hierarchical features and capture temporal relationships in financial markets\cite{bao2017deep}.

However, single machine learning methods face issues like overfitting and may struggle to fit all scenarios in financial markets\cite{de2020machine} perfectly. Therefore, researchers tend to blend different models, leveraging their respective strengths\cite{li2012complex}, such as combining statistical methods with LSTM to utilize their advantages in both linear and nonlinear modeling\cite{bukhari2020fractional}. In terms of evaluation metrics, regression, and classification metrics are widely used\cite{tang2022survey}. Machine learning provides practical modeling tools for financial time series prediction, but challenges related to sample size, prediction performance, and other factors require further research\cite{hsu2016bridging}. Future research should focus on the further application of deep learning and hybrid methods.

\subsection{Gramian Angular Field for Financial Time Series Forecasting}
In recent years, GAF has been widely used to convert time series data into images to leverage the advantages of CNNs in image processing for time series modeling and prediction. In the financial field, GAF plays an influential role in encoding financial time series data by ingeniously mapping the angular information in time series into images. At the same time, CNNs can efficiently extract spatial features implied in GAF images. For example, Barra \emph{et al.} constructed a CNN ensemble model based on the S\&P 500 futures, and they found that this method can not only improve the prediction accuracy but also obtain higher returns, which demonstrated the value of the GAF-CNN framework in financial forecasting\cite{barra2020deep}. Chen and Tsai applied GAF-CNN to identify candlestick patterns of stocks, and its performance achieved over 90\% accuracy on both simulated and accurate data, showcasing the superior identification capability of this method\cite{chen2020encoding}. Lee and Ko considered multiple variables to construct GAF images. They applied CNNs to predict fund prices, and the results showed that the multi-variable model had significantly higher cumulative returns than the single-variable model, revealing the benefits of incorporating multiple influencing factors\cite{lee2019fund}. Wu \emph{et al.} used SIFT features of GAF images to cluster financial time series and found that this method can effectively detect anomalies in financial sequences, providing new perspectives for financial monitoring\cite{wu2023imaging}. Yang et al. proposed a financial time series label correction method based on meta-learning and contrastive learning. They achieved results superior to multiple baselines, providing new ideas for automatically generating accurate labels for financial time series\cite{yang2023meta}. 

In summary, GAF is an efficient way of time series imaging, and its combination with CNNs has demonstrated great potential in enhancing the representation and prediction capabilities of financial time series data.

\subsection{Quantum Computing for Finance}
\label{sec:others}
With the development of quantum computing hardware and algorithms, quantum computing holds vast prospects for applications in the financial sector, particularly in machine learning. Herman \emph{et al.} highlighted that quantum machine learning models, such as quantum neural networks, can be applied to financial prediction tasks like stock price forecasting\cite{herman2023quantum}. This is primarily based on the inherent advantages of quantum computing in sampling and representing complex distributions\cite{herman2023quantum}. Furthermore, quantum computing finds utility in financial domains for tasks such as stochastic modeling\cite{stamatopoulos2020option,ramos2021quantum,kubo2021variational}, optimization\cite{mugel2021hybrid,rebentrost2022quantum,he2023alignment}, dynamic programming\cite{ambainis2019quantum}, demonstrating significant speed advantages when compared to classical algorithms. For instance, Quantum Monte Carlo Integration(QMCI) can achieve quadratic acceleration in pricing and risk analysis tasks. Some quantum heuristic algorithms, like the quantum variational algorithms, have also shown partial performance improvements in combinatorial optimization problems\cite{liu2022layer}. However, these quantum algorithms face challenges like model loading and sampling complexity. For practical speed advantages to be realized, there is a need to reduce resource requirements further and enhance quantum hardware\cite{herman2023quantum,babbush2021focus}.

In conclusion, quantum computing is poised to address particular computational challenges in the financial domain soon, potentially creating a revolutionary impact on financial practices.

\section{Methodology}
\subsection{Time Series Data Segment}
\begin{figure}[H]
\centering
\includegraphics[width=13.5 cm]{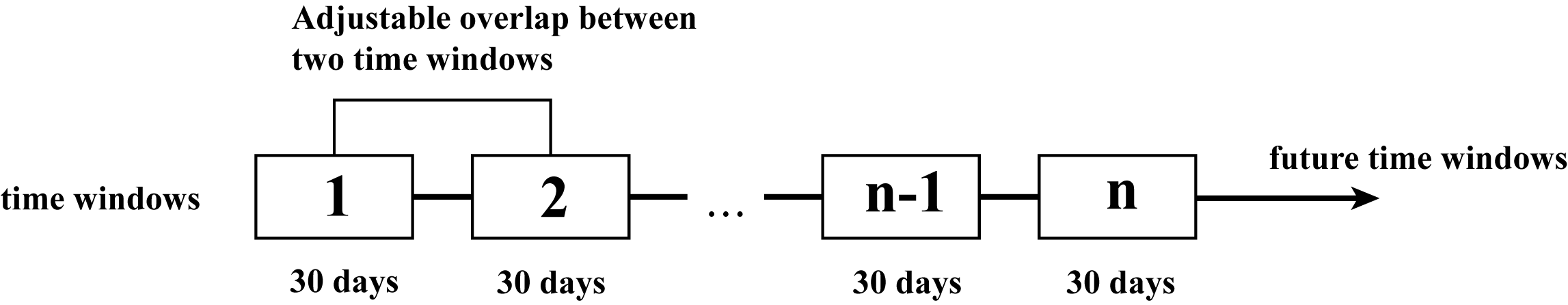}
\caption{Sliced time series window with some adjustable overlap between windows.\label{fig1}}
\end{figure}

The overarching objective of our research centers on harnessing the information encapsulated within historical time windows to forecast the interval returns for the subsequent adjacent time window. This task, in its very essence, demands a nuanced approach, and our chosen methodology is to deploy a CNN with a tailored architecture to achieve this prediction. Figure \ref{fig1} shows that our image preparation process involves segmenting time series data and implementing a sliding window computation.

A balance in image content is pivotal. Small pixel dimensions could result in undersampling of the content, while expansive ones overburden the image, complicating the CNN's learning process and feature extraction. To navigate this, we employ a rolling time window that spans 30 days of returns, yielding a 30×30 pixel image.

The choice of a 10-day stride coupled with a 20-day overlap between successive windows isn't arbitrary. This configuration ensures that there's a substantial overlap between the historical data of one window and the subsequent one. Such a design is imperative for two reasons. Firstly, it facilitates CNN in recognizing patterns and trends that evolve over longer periods, ensuring that the transitional data between windows isn't lost. Secondly, by emphasizing this overlap, the model is better equipped to infer and predict the interval returns of the next time window, given that it maintains continuity with the preceding data. This overlapping mechanism, akin to a one-dimensional convolution, serves as a bridge connecting adjacent time windows, making the prediction more holistic and grounded in historical context.

\unskip

\subsection{Data Visualization Method}

\subsubsection{Gramian Angular Field}\label{sub3.2.1}
Wang and Oates introduced a time-series encoding method called GAF\cite{wang2015imaging}. GAF achieves this by normalizing the time series to the [-1,1] or [0,1] interval and representing its values as angular cosines in polar coordinates, with timestamps as radii. This transformation converts the time series into a polar coordinate representation. Using this foundation, we define the Gramian Angular Summation Field (GASF) and Gramian Angular Difference Field (GADF) matrices. These matrices capture the correlations between values in the time series at different time spans through cosine and sine operations, forming an image representation of the temporal signal. This GAF image representation method provides a certain degree of data compression while preserving the temporal dependencies within the time series.

Given a segmented time series \(X = \{x_1, x_2, ..., x_n\}\) of length \(n\), the first step is to perform normalization so that the values fall within the range of [0,1] or [-1,1]:

\begin{equation}\label{eq1}
\tilde{x}_i = \frac{x_i - \min(X)}{\max(X) - \min(X)}    
\end{equation}

\begin{equation}\label{eq2}
\tilde{x}_i = \frac{(x_i - \max(X) + (x_i - \min(X))}{\max(X) - \min(X)}    
\end{equation}

Next, the series is transformed into polar coordinates:

\begin{numcases}{}
\phi = \arccos(\tilde{x}_i), & $\tilde{x}_i \in \tilde{X}$ \label{eq3} \\
r = \frac{t_i}{N}, & $t_i \in N$ \label{eq4}
\end{numcases}

Here, \(t_i\) represents the time step, and \(N\) is the normalization factor.

Based on the polar coordinate representation, the Gramian Angular Field (GAF) is defined to represent the correlation of time series at different time intervals. Defining the inner product \(\langle x, y \rangle = x \cdot y - \sqrt{1 - x^2}\sqrt{1 - y^2}\), the Gramian Angular Summation Field (GASF) and Gramian Angular Difference Field (GADF) are defined as follows:

\begin{equation}\label{eq5}
GASF = \cos(\phi_i + \phi_j)   
\end{equation}

\begin{equation}\label{eq6}
GADF = \sin(\phi_i - \phi_j)    
\end{equation}

GASF can be represented through matrix operations:

\begin{equation}
GASF = \tilde{X}' \tilde{X} - \sqrt{I - \tilde{X}^2}' \sqrt{I - \tilde{X}^2}    
\end{equation}

GADF can be represented as:

\begin{equation}
GADF = \sqrt{I - \tilde{X}^2}' \tilde{X} - \tilde{X}' \sqrt{I - \tilde{X}^2}  
\end{equation}
where \(I\) is an all-ones vector.

After these transformations, the time series is represented as GAF images, and then image processing methods can be applied for time series analysis.

\subsubsection{Quantum Gramian Angular Field}

In the preceding section, the classical GAF encodes time series by converting them into angular cosines in polar coordinates. This method includes GASF and GADF, offering a unique image representation of time series data. Building on this classical approach, the Quantum Gramian Angular Field (QGAF) takes a quantum-inspired route. Instead of following traditional computational steps, QGAF efficiently encodes the trigonometric relationships essential for GAF into quantum gate parameters. This strategy, similar to the approach in Quantum Phase Estimation (QPE) algorithms\cite{kitaev1995quantum, nielsen2010quantum}, allows the desired values to be embedded in the amplitude of quantum states. These values are then directly retrieved by performing quantum measurements, eliminating the need for normalization and calculations of arccosine. This approach highlights the potential of using quantum mechanics to simplify complex computations, especially when analyzing stock return data.

For a segmented time series depicted as \( X=\{x_1,x_2,...,x_n\} \) with a length \( n \), we can extrapolate two sequential data points: \( a=x_i \) and \( b=x_j \). Using a solitary qubit, \( q_0 \), the quantum states of these angles find representation through the \( Ry \) rotation gate. To derive the Quantum Gramian Angular Summation Field (QGASF), which is the quantum alternative for GASF, we first need to compute \( \cos(a+b) \). Using the \( R_y \) rotation gate.

\begin{equation}
Ry(\theta) = 
\begin{bmatrix}
\cos(\frac{\theta}{2}) & -\sin(\frac{\theta}{2}) \\
\sin(\frac{\theta}{2}) & \cos(\frac{\theta}{2})
\end{bmatrix}   
\end{equation}

and starting with \( |0\rangle = \begin{bmatrix} 1 \\ 0 \end{bmatrix} \), we first apply \( Ry(2a) \) to get:

\begin{equation}
Ry(2a) |0\rangle = 
\begin{bmatrix}
\cos(a) \\
\sin(a)
\end{bmatrix}
\end{equation}

Subsequent application of \( Ry(2b) \) yields:

\begin{equation}
Ry(2b) 
\begin{bmatrix}
\cos(a) \\
\sin(a)
\end{bmatrix}
=
\begin{bmatrix}
\cos(a)\cos(b) - \sin(a)\sin(b) \\
\cos(a)\sin(b) + \sin(a)\cos(b)
\end{bmatrix}
=
\begin{bmatrix}
\cos(a+b) \\
\sin(a+b)
\end{bmatrix}
\end{equation}

The probability $p$ of measuring the qubit in the $|0\rangle$ state after rotation translates to:

\begin{equation}
p = \cos^2(a+b) \implies \cos(a+b) =\pm\sqrt{p}
\end{equation}

On the other hand, for the computation of \( \sin(a-b) \), which is essentially the computation for the Quantum Gramian Angular Difference Field (QGADF), we start with the same initial state \( |0\rangle \) as the starting and after applying \( Ry(2a) \), we obtain:

\begin{equation}
Ry(2a) |0\rangle = 
\begin{bmatrix}
\cos(a) \\
\sin(a)
\end{bmatrix}
\end{equation}

A subsequent application of \( Ry(-2b) \) then provides:

\begin{equation}
Ry(-2b) 
\begin{bmatrix}
\cos(a) \\
\sin(a)
\end{bmatrix}
=
\begin{bmatrix}
\cos(a-b) \\
\sin(a-b)
\end{bmatrix}
\end{equation}

Thus, the probability \( p' \) of measuring the qubit in the \( |1\rangle \) state translates to:

\begin{equation}
p' = \sin^2(a-b) \implies \sin(a-b) =\pm\sqrt{p'}
\end{equation}

In theory, adopting a single qubit complemented by a series of $R_y$ rotations furnishes a quantum alternative for calculating the summation and difference of angles via their cosine and sine representations. This approach not only exemplifies the capabilities of quantum computation in traditional trigonometric operations but also heralds an innovative frontier in quantum financial analysis.

\begin{figure}[H]
\centering
\includegraphics[width=10.5 cm]{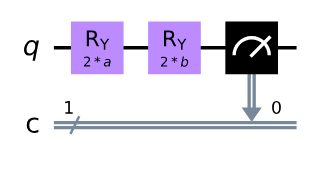}
\caption{The schematic diagram of the quantum circuit for QGASF.\label{fig2}}
\end{figure}

We present the associated quantum circuits to offer a clearer understanding of QGAF. Figure \ref{fig2} demonstrates the quantum circuit designed explicitly for QGASF. Its goal is to compute \( \cos(a + b) \). The circuit works on a single qubit initialized in the \( |0\rangle \) state. It involves two consecutive \( R_y \) rotations: the first with an angle of \( 2a \), followed by a second one at \( 2b \). After measuring the qubit, the probability of finding it in the \( |0\rangle \) state provides an approximation for \( \cos(a + b) \). 

\begin{figure}[H]
\centering
\includegraphics[width=10.5 cm]{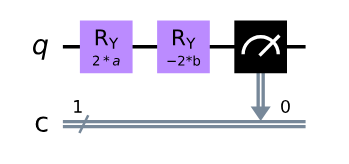}
\caption{The schematic diagram of the quantum circuit for QGADF.\label{fig3}}
\end{figure}

Similarly, Figure \ref{fig3} shows the quantum circuit for QGADF, which calculates \( \sin(a - b) \). This circuit also uses a single qubit in the \( |0\rangle \) state but differs in the second rotation, which is set to \( -2b \). The measurement probability for the \( |1\rangle \) state gives an approximation of \( \sin(a - b) \).

\unskip

\unskip

By running these circuits multiple times and averaging the results, we can improve the accuracy of our calculations. These circuits help to identify relationships in time series data and are essential in creating QGAF images. It's worth noting that most GAF-related studies use the GASF method\cite{hsu2021using, chen2020encoding,wu2023imaging,yang2023meta}. Therefore, in our stock time series data analysis, we chose QGASF as our method to convert one-dimensional time series data into two-dimensional images, and we used CNNs for further training.

\subsection{Training processes of two-dimensional images based on CNNs}

\begin{figure}[H]
\centering
\includegraphics[width=18cm]{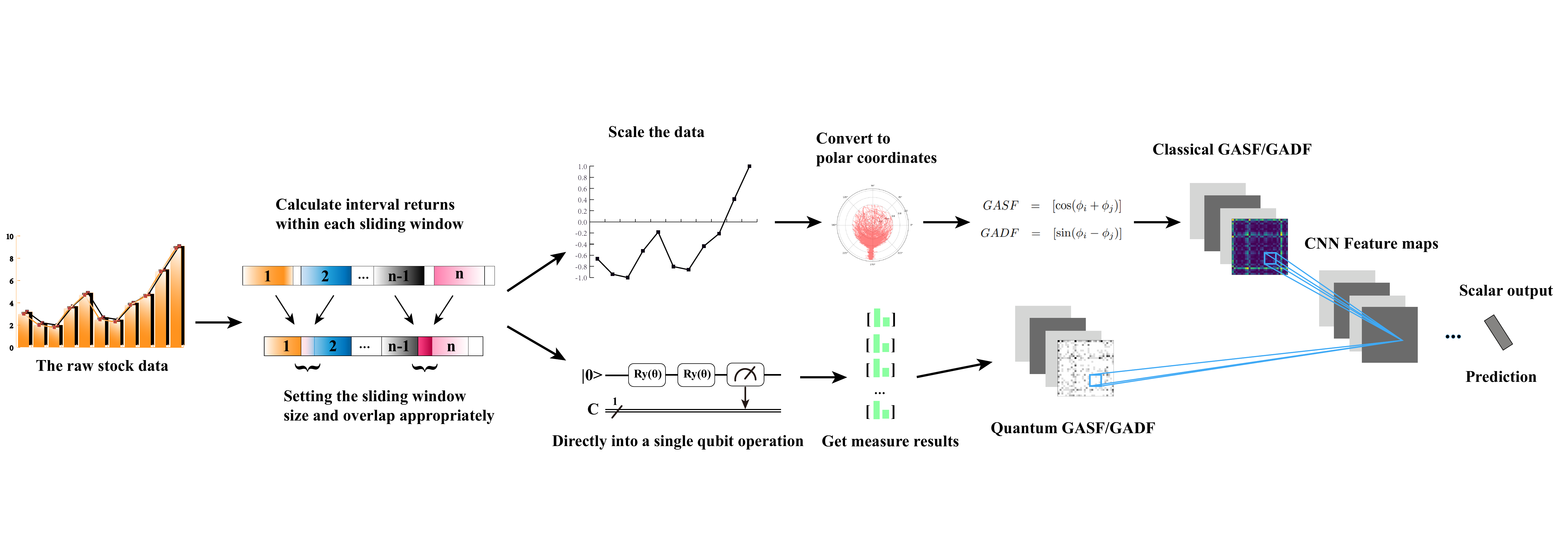}
\caption{Overall framework for training a CNN.\label{fig4}}
\end{figure}

In our investigation of QGASF images, a vital step involves the utilization of CNNs for end-to-end training. This training approach is particularly suitable for time series classification. By transitioning from one-dimensional time-domain data to two-dimensional images, this methodology effectively harnesses the inherent strengths of CNNs in image processing, thus potentially enhancing the predictive accuracy for time series data classification. This research follows the workflow shown in Figure \ref{fig4}.

\unskip

\begin{figure}[H]
\centering
\includegraphics[width=18cm]{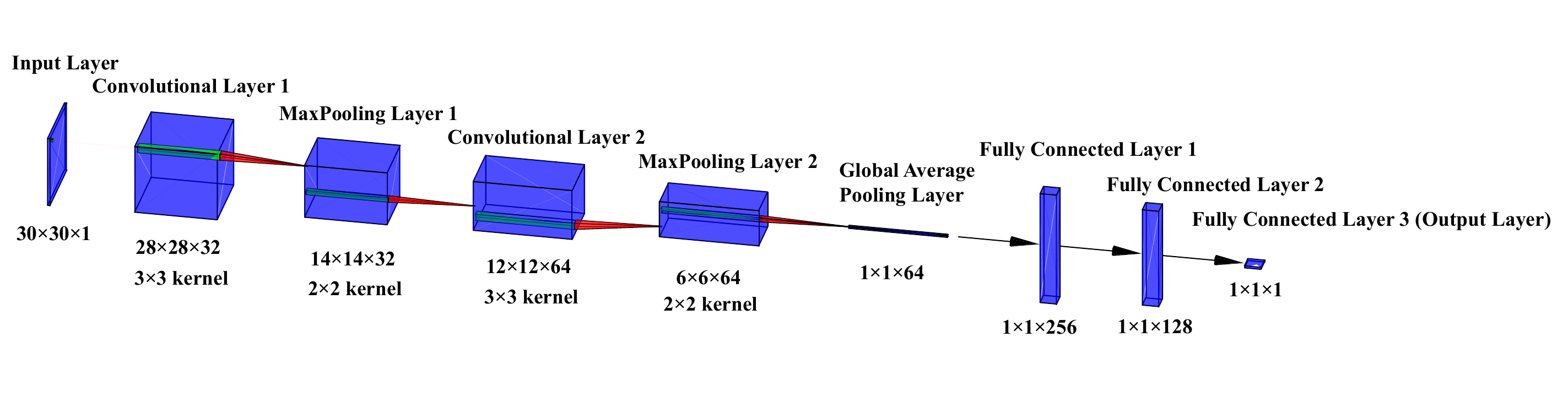}
\caption{Architecture of the Predictive CNN for Time Series Analysis.\label{fig5}}
\end{figure}

The network architecture, developed using PyTorch, is central to the process. It is designed to handle single-channel (grayscale) 30x30 pixel images. These images undergo transformations, including grayscaling and normalizing pixel values to the 0-1 range. The core CNN architecture, depicted in Figure \ref{fig5}, comprises convolutional layers, a global average pooling layer, and several fully connected layers. Specifically, the convolutional segment has two layers, each followed by a ReLU activation and max-pooling. The fully connected segment is tailored for regression, producing a single output value.

Model training is systematic, encompassing model initialization, loss function specification, and optimizer selection. The loss function is either MSE or MAE, and the chosen optimizer is Adam. The training process spans multiple epochs, with the model undergoing forward and backward passes, loss-based optimization, and validation set evaluation. Post-training, the model’s state dictionary is preserved for subsequent use. This stored model facilitates quick loading for predictions, supporting real-time assessments and implementations.

\unskip

\section{Experiments}
\subsection{Datasets}
We used daily return time series of multiple stocks from the China A-share market (600030. SH, 600036. SH, 600519. SH, 601166. SH, and 601318. SH), the Hong Kong stock market (0005. HK, 0700. HK, 0939. HK, 0941. HK, and 1299. HK), and the U.S. stock market (Boeing, Dow Chemical, IBM, JPMorgan Chase, and Procter \& Gamble). These datasets cover stock markets in three countries and regions: China, Hong Kong, and the United States. Representative blue-chip stocks with significant trading volumes were selected as samples, ensuring a specific scale and representativeness that can effectively reflect the statistical characteristics of financial time series data.

\subsection{Experimental Settings}

\subsubsection{Data Collection and Preprocessing}
The data acquisition was done using Python libraries: the yfinance library from Yahoo Finance for U.S. stocks and the Tushare library for both the China A-share and the Hong Kong stock markets. Notably, all selected stocks have data post-2000, though stocks introduced later might exhibit shorter data durations.

Our preprocessing steps were meticulous, ensuring the time series data was void of inconsistencies. This process encompassed data cleaning and addressing missing values. However, we did not address outliers, as stock prices can exhibit significant fluctuations for various reasons, such as news events, company announcements, or market crashes. While these fluctuations might statistically be identified as outliers, they are, in fact, genuine market reactions. For continuous missing data, we employed forward-filling techniques. Additionally, the five-day moving average method was used to refine the treatment of missing values further. An essential preprocessing step was the computation of daily returns. The daily return \( R_t \) for a given stock on day \( t \) is given by:

\begin{equation}
R_t = \frac{{\text{Closing Price}_t - \text{Closing Price}_{t-1}}}{{\text{Closing Price}_{t-1}}}
\label{eq16}
\end{equation}

Where \( \text{Closing Price}_t \) is the closing price on the day \( t \) and \( \text{Closing Price}_{t-1} \) is the closing price on the previous day.

\begin{figure}[H]
\centering
\includegraphics[width=15.5cm]{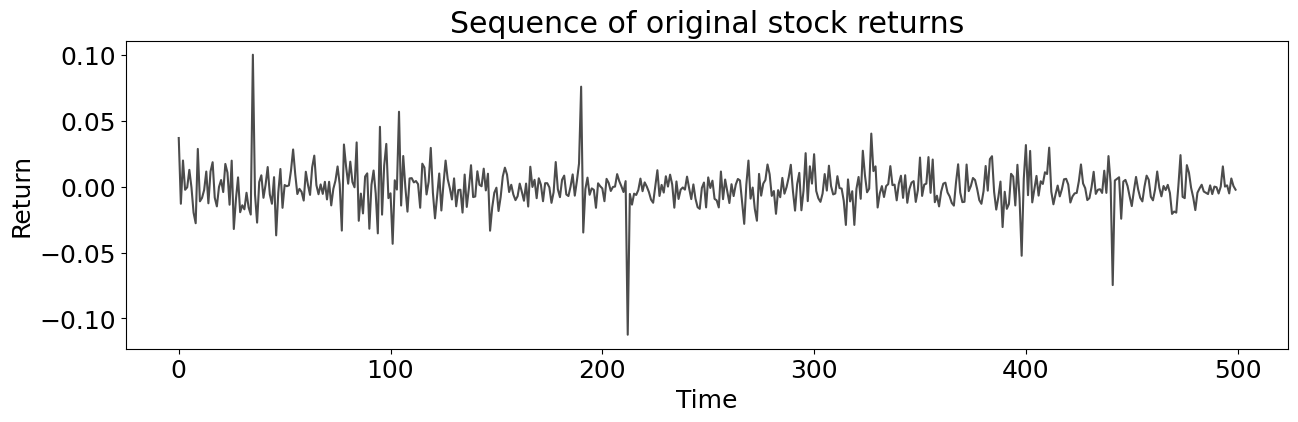}
\caption{Time-series chart of Moutai's return over the previous 500 trading days.\label{fig6}}
\end{figure}

Figure \ref{fig6} presents a time-series chart of the daily returns for Moutai stock over the previous 500 trading days. The daily return $R_t$ is calculated based on equation \ref{eq16}. As depicted in the figure \ref{fig6}, the daily returns of Moutai stock have experienced significant fluctuations during this period. These fluctuations reflect the market performance of the stock, potentially influenced by various factors such as market conditions, company news, and macroeconomic data. This figure demonstrates one of the empirical datasets we used to evaluate the effectiveness of QGASF and CNNs in predicting stock returns.

\subsubsection{Image Preparation for GASF and QGASF}
Our next step was the generation of both classical GASF and QGASF images. Using the methods previously described, we produced 15 folders each for the classical GASF and QGASF images. The image preparation process required careful segmenting of time series data, determining sliding windows, and setting the pixel dimensions for each image. We aimed to avoid over-complicated images that hinder CNN learning, hence settling on a 30-day return period as a rolling window, resulting in 30x30 pixel images. This approach is analogous to one-dimensional convolution. We adopted a 10-day stride and a 20-day overlap for successive windows to maintain continuity between operations across windows.

The cumulative return for a time window, starting from day 1 (\( t_1 \)) to day 30 (\( t_{30} \)), is given by:

\begin{equation}
\text{Interval Return} = \prod_{i=1}^{30} (1 + R_{t_i})
\end{equation}

Where \( R_{t_i} \) is the daily return on day \( t_i \).

\subsubsection{CNN Framework and Evaluation}
After preprocessing, the two-dimensional image data for both the classical GASF and QGASF were prepared. On average, most stocks generated approximately 500 images. Due to their extended post-2000 history, U.S. stocks yielded up to 595 images. Before CNN training, each image was systematically paired with its respective label, indicating the cumulative return for that specific period.

We chose Adam as our optimization algorithm for the training process and employed MSE and MAE as loss functions.
We trained the model over 100 epochs, covering the whole dataset in each epoch. Based on our observations, 100 epochs
provided satisfactory model convergence, considering the image dataset’s manageable size and the simplicity of pixel
data. A cross-validation strategy was implemented, comprising five iterations of training and validation. Each iteration
used 80\% of the data for training and the remaining 20\% for validation. We rigorously assessed the enhancement
capabilities and image quality of both QGASF and GASF, alternating between the training and validation phases. Both
MAE and MSE were utilized as loss metrics, which facilitated the derivation of experimental results for GASF and
QGASF. All computations were conducted on an NVIDIA A100 GPU server.

\subsection{Results}

\begin{figure}[H]
\centering
\includegraphics[width=10.5 cm]{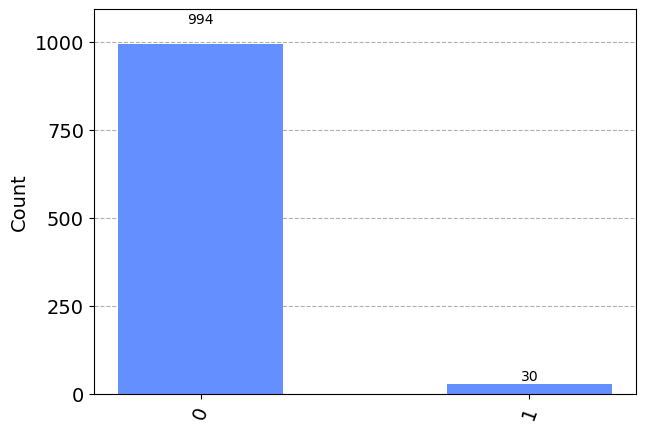}
\caption{Quantum measurement results for given $a=0.05$, $b=0.1$.\label{fig7}}
\end{figure}

Quantum measurement plays a pivotal role in quantum computing. To obtain the QGASF images suitable for CNN training, we first need to measure the quantum circuit of QGASF. Figure \ref{fig7} displays the measurement results for the quantum circuit given $a=0.05$ and $b=0.1$. As can be inferred from the figure, out of 1024 measurements on qubit $q$, 994 measures resulted in the $|0\rangle$ state, while 30 measurements were in the $|1\rangle$ state. This suggests that the estimated value of $cos(a+b)$ is approximately 0.9852426731521529.
In comparison, the classical computation provides a value of 0.988771077936042. While these results exhibit proximity, it's noteworthy that quantum methodologies are susceptible to specific inherent random measurement errors. Such nuances manifest as subtle noise, crucial for enhancing training robustness when transcribing computations into imagery on a pixel-by-pixel foundation.

\unskip

\begin{figure}[H]
\centering
\includegraphics[width=16.5cm]{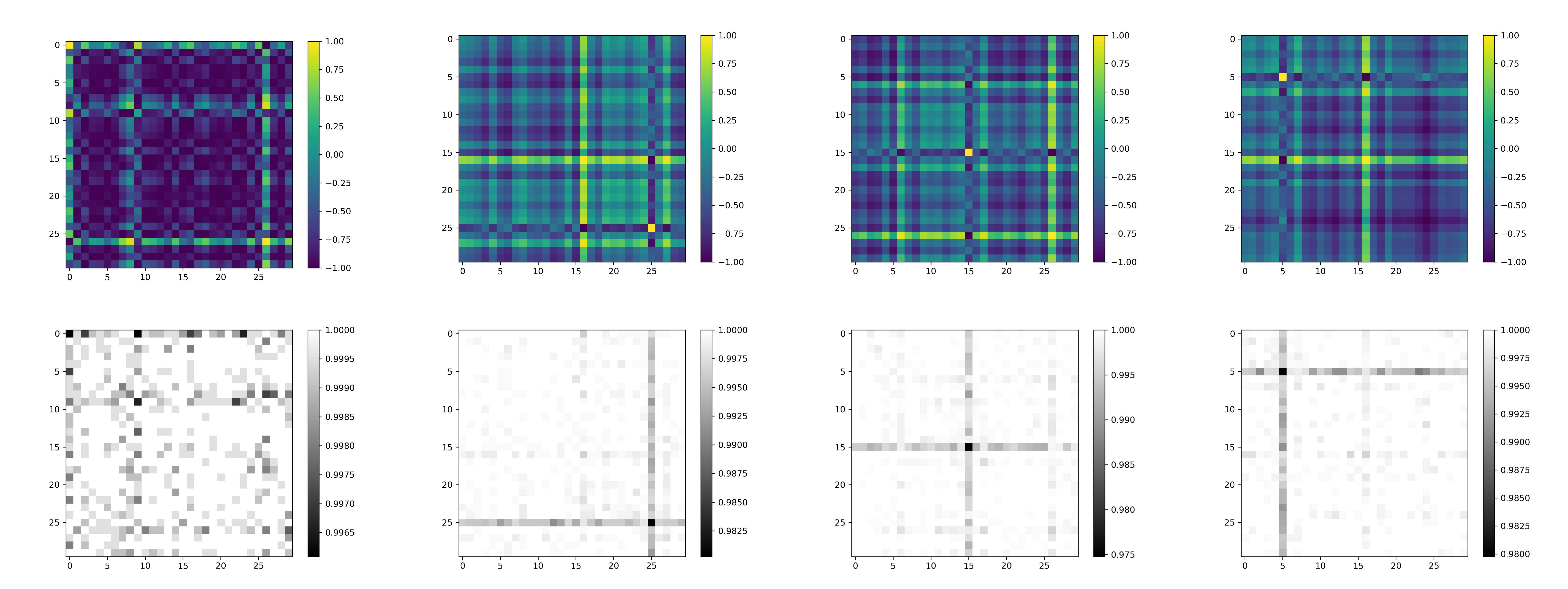}
\caption{Comparison of the performance between classical GASF and QGASF for Moutai single share (600519.SH), with the top four sub-figures showing partial schematics of classical GASF and the bottom four sub-figures showing partial schematics of QGASF.\label{fig8}}
\end{figure}

As aforementioned, QGASF obviates the need for data normalization and the arccos computation relative to GASF (refer to Appendix \ref{Appendix B} for details on classical GASF). Figure \ref{fig8} provides a comparative visualization between the classical GASF and the QGASF for Moutai single share (600519. SH). To distinguish from the GASF images, we set the color parameter of the QGASF images to gray. Both depictions exhibit similarities in image structure, pattern distribution, and specific details, reflecting their efficacy in capturing fundamental characteristics of time series data. Notably, the grayscale images of QGASF capture distinct features from the classical GASF and introduce random noise inherent to quantum measurements. This stochastic noise augments the robustness and generalization capabilities of CNN training, ensuring consistent predictive accuracy across varied scenarios. Consequently, this suggests that QGASF, in the context of stock time series forecasting, may outperform GASF, potentially offering enhanced insights and cues for stock return predictions.

To quantitatively investigate the differences in stock return prediction performance between QGASF and GASF methods, we selected five blue-chip stocks from three different stock markets: the China A-share market, the Hong Kong stock market, and the US stock market, to serve as our training and testing datasets. We chose MAE and MSE as metrics to evaluate the model's predictive performance. The smaller the MAE and MSE values on the test set, the better the predictive efficacy of the model.

\begin{figure}
\centering
\includegraphics[width=15.5cm]
{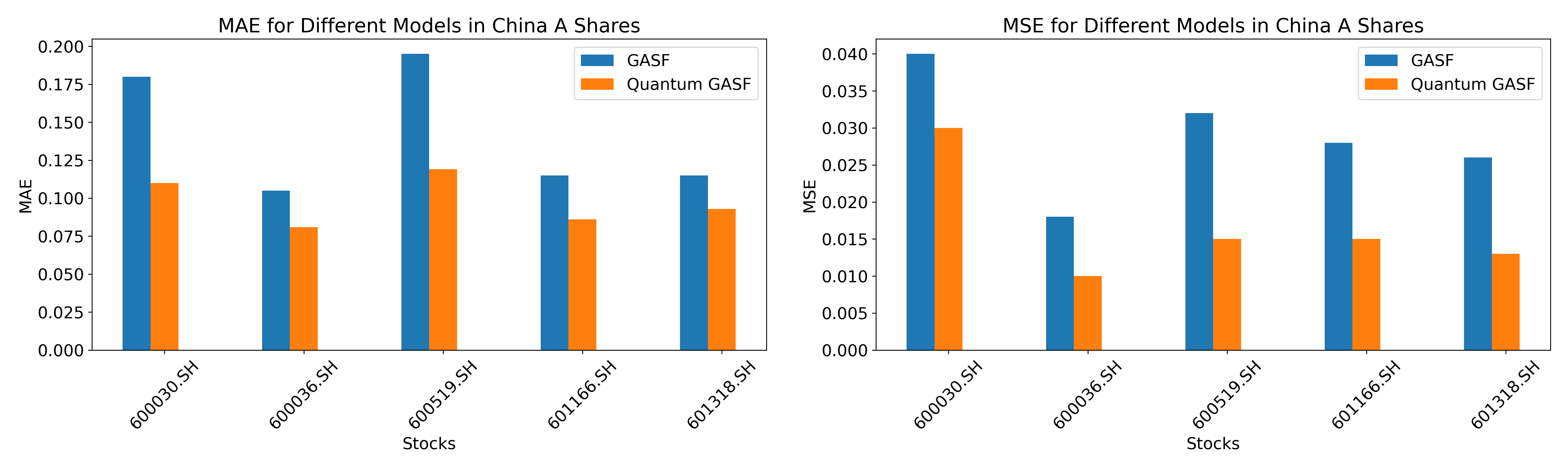}
\caption{MAE(left) and MSE(right) for different models in the China A-share market.\label{fig9}}
\end{figure}

\begin{figure}
\centering
\includegraphics[width=15.5cm]
{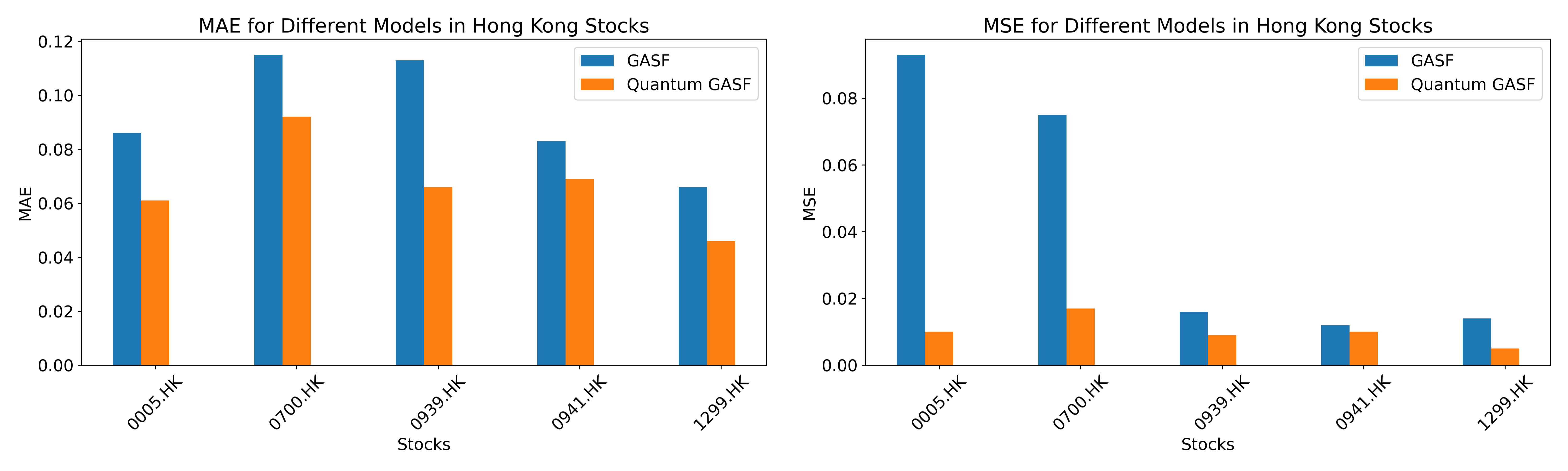}
\caption{MAE(left) and MSE(right) for different models in the Hong Kong stock market.\label{fig10}}
\end{figure}

\begin{figure}
\centering
\includegraphics[width=15.5cm]{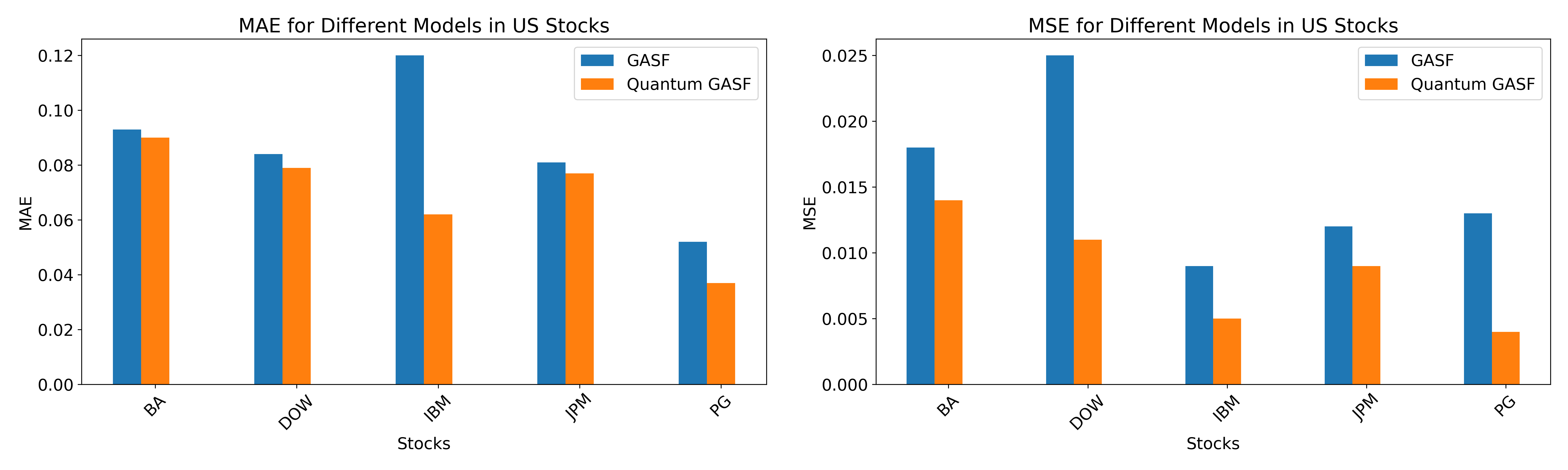}
\caption{MAE(left) and MSE(right) for different models in the US stock market.\label{fig11}}
\end{figure}

Figures \ref{fig9}-\ref{fig11} show the predictive performance of the trained models in the China A-share market, the Hong Kong stock market, and the US stock market, respectively. The QGASF method outperforms the traditional GASF. Figure \ref{fig9} reveals that in the China A-share market, QGASF reduces the MAE and MSE errors by an average of 29\% and 44\%, respectively. As depicted in Figure \ref{fig10}, these two metrics in the Hong Kong stock market decrease on average by 28\%  and 58\%, respectively. Notably, stock 0700. HK experienced the most significant reduction in MSE, reaching 89\%. In the US stock market (Figure \ref{fig11}), while the decrease in MAE and MSE for QGASF might not be as pronounced for certain stocks compared to other markets, overall, it still demonstrates superior predictive capacity, with average reductions in MAE and MSE of 18\% and 43\%, respectively. In total, MAE decreased by an average of 25\%, and MSE decreased by an average of 48\% in these three stock markets. This remarkable reduction in error can be attributed to several critical advantages of QGASF: it doesn't require normalization of the data, which, in the specific context of stock return prediction, preserves the concentration of the original data without dispersing it; the noise in quantum measurements enhances the generalizability of the model; and QGASF does not involve an inverse cosine operation, thus effectively retaining the sign information in the data, crucial for determining the rise or fall of stock returns. Collectively, these factors make QGASF a more accurate and stable method for predicting stock returns.

The results of validation across stock markets in three different countries and regions indicate that the QGASF method significantly enhances the accuracy of time series predictions compared to the classical GASF method. This validates that combining quantum computing with deep learning yields improved expressive power and predictive performance. This approach holds promising applications in financial time series forecasting, and its generalization performance can be further validated in various real-world scenarios.

\begin{figure}[H]
\centering
\includegraphics[width=15.5cm]{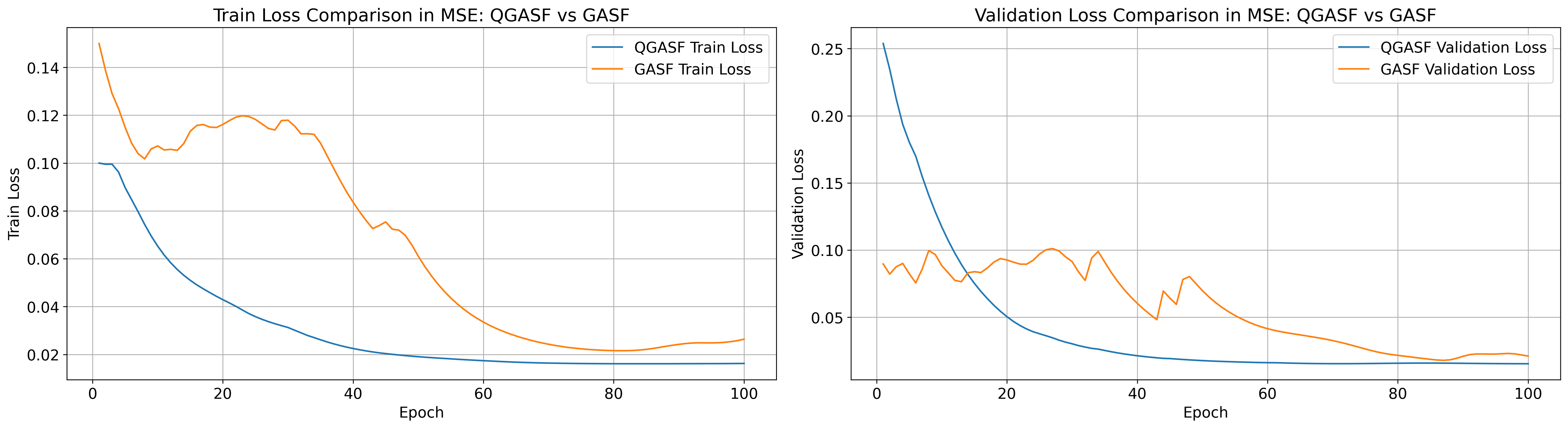}
\caption{MSE on Moutai single share data(600519.SH). Left panel: train loss comparison between QGASF and GASF in MSE. Right panel: validation loss comparison between QGASF and GASF in MSE.\label{fig12}}
\end{figure}

\begin{figure}[H]
\centering
\includegraphics[width=15.5cm]{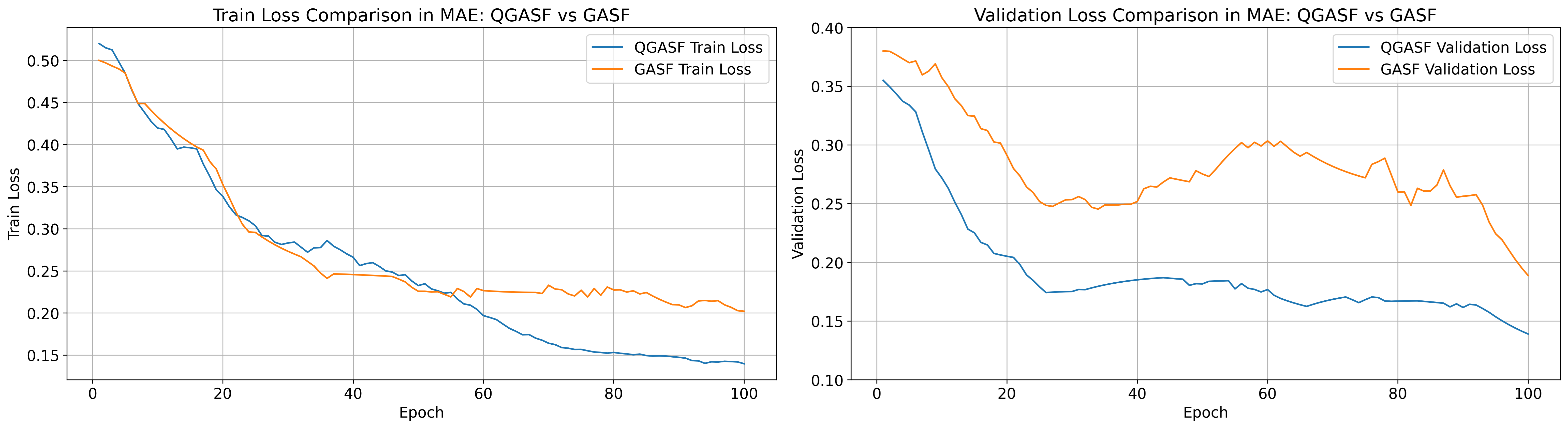}
\caption{MAE on Moutai single share data(600519.SH). Left panel: train loss comparison between QGASF and GASF in MAE. Right panel: validation loss comparison between QGASF and GASF in MAE.\label{fig13}}
\end{figure}

The performance differences between QGASF and GASF are also clearly evidenced in the MSE and MAE loss curves, as depicted in Figures \ref{fig12} and \ref{fig13}. On the MSE curve for the training set, the loss of QGASF descends more rapidly than that of GASF. Around the 10th epoch, QGASF's loss begins to significantly underperform GASF, while GASF's loss enters a period of oscillation, indicating superior early-stage learning performance by QGASF. QGASF begins to converge around the 50th epoch, whereas GASF starts its convergence near the 80th epoch, showcasing a noticeably faster convergence time for QGASF. Moreover, throughout the learning process, QGASF consistently surpasses GASF. The MSE loss curve for the validation set reveals that although QGASF's MSE curve initially trails GASF's, its rapid decline soon leads to a cross-over. Post this intersection, QGASF's loss remains lower. It starts converging around the 50th epoch, underscoring the stability and superiority of QGASF images throughout the training process, further attesting to its exceptional generalization capabilities.

The MAE loss curves further corroborate this assessment. Even though QGASF's loss doesn't always decrease faster than GASF's in the training set, its ultimate loss value is distinctly lower, with their early-stage declines roughly parallel. It's evident that after the 60th epoch, QGASF's converged value is significantly lower than GASF's, highlighting its enhanced stability and robustness. Notably, the MAE loss curve for the validation set demonstrates that QGASF maintains a loss lower than GASF's throughout nearly all epochs, further testifying to its outstanding generalization capability. Analysis of the loss curves from training and validation sets also suggests that our CNN model doesn't suffer from overfitting. 

By comparing the four sets of loss curves, it becomes clear that QGASF grayscale images offer many advantages, such as high quality, distinct features, rich information, ease of training, and excellent generalization capabilities. This outcome aligns with our expectations, given that the measurement error in quantum circuits introduces innate random noise to the images generated by QGASF, fortifying the model's robustness and generalizability. Moreover, since QGASF doesn't employ the inverse cosine mapping found in GASF, it preserves the positive and negative label information, enabling effective prediction of stock return trends. The streamlined computation process renders QGASF image features more lucid, facilitating CNN's learning.

\section{Conclusions}

This paper introduces a novel time series forecasting method called QGAF. This method combines classical GAF with quantum computing technology to produce superior time series forecasting results compared to classical forms. The article begins by providing an overview of the background and significance of time series forecasting tasks and the pros and cons of the classical GAF method. It then provides a detailed description of the QGAF methodology, including quantum circuit design and measurement result interpretation. Finally, an end-to-end deep learning framework is designed to train QGAF images using CNNs for time series classification.

Comparative experiments conducted on different stock market datasets demonstrate that the QGAF method significantly enhances the effectiveness of time series classification compared to approaches solely reliant on time series or classical image-based methods. This underscores the feasibility and efficacy of combining quantum computing with deep learning algorithms, offering an instructive example for similar problem domains.

The QGAF method has shown significant potential in time series forecasting tasks. However, the fusion of two budding areas offers many research avenues to delve into. One critical aspect is validating the method's generalization capabilities across a broader spectrum of real-world datasets. This would provide insights into its applicability and robustness. Additionally, exploring the synergy of deep learning models other than CNNs with QGAF representations is crucial. Another pivotal area is the development of strategies to quantize conventional deep models, facilitating complete quantum training. We are confident that continued research in these domains will make QGAF technology a formidable asset for time series analysis. Our optimism also extends to the broader realm of quantum computing in machine learning, and we eagerly await future groundbreaking innovations.

\bibliographystyle{plainnat}
\bibliography{references}  

\titleformat{\section}[block]{\normalfont\Large\bfseries}{Appendix \Alph{section}:}{1em}{}

\clearpage
\appendix

\renewcommand{\thefigure}{\thesection.\arabic{figure}}
\renewcommand{\thetable}{\thesection.\arabic{table}}

\makeatletter
\@addtoreset{figure}{section}
\@addtoreset{table}{section}
\makeatother

\section{Pseudo Codes}

The QGASF Images Generation Algorithm \ref{algorithm1} aims to generate QGASF images from an array of stock returns. The algorithm is fed an array of stock returns, a pre-determined window size for segmenting time, and a stride value for navigating the returns array. The final output consists of a list of QGASF images.

Initially, the algorithm initializes an empty list, $quantum\_gasfs$. It then traverses the returns array with a defined stride, segmenting the returns based on the input window size. These segments form the basis for generating the QGASF images.

For each segment, a 2D array, $quantum\_gasf$, is initialized with dimensions equivalent to the window size. The algorithm processes each pair of returns in the segment to create a quantum circuit. Each circuit is initialized with a single qubit in the quantum register and a single bit in the classical register. $R_Y$ gates are applied to the qubit using the return values as angles. After these quantum operations, a measurement is taken and simulated. The simulation results populate the $quantum\_gasf$ array, which, upon completion, gets appended to the $quantum\_gasfs$ list. The algorithm concludes once all segments have been processed in this manner.

\begin{algorithm}[H]
\caption{QGASF Images Generation}
\label{algorithm1}
\begin{algorithmic}[1]
\STATE {\bfseries Input:}
\STATE \quad returns: Array of stock returns
\STATE \quad window\_size: Size of the time window for segmentation
\STATE \quad stride: Stride for moving the time window
\STATE {\bfseries Output:}

\STATE \quad $quantum\_gasfs$: List of QGASF images

\STATE Initialize empty list $quantum\_gasfs$
\FOR{$i\gets 0$ {\bfseries to} $len(returns)$ {\bfseries step} $stride$}
\STATE $segment \gets returns[i:i + window\_size]$

\STATE Append $segment$ to $segments$
\ENDFOR

\FOR{each $segment$ in $segments$}
\STATE Initialize 2D array $quantum\_gasf$ with shape $(window\_size, window\_size)$

\FOR{$i\gets 0$ {\bfseries to} $window\_size$}

\FOR{$j\gets 0$ {\bfseries to} $window\_size$}
\STATE $a \gets segment[i]$
\STATE $b \gets segment[j]$

\STATE {\bfseries // Construct quantum circuit}
\STATE Preparing the quantum initial state $|0\rangle$
\STATE Initialize quantum register $qreg$ with 1 qubit
\STATE Initialize classical register $creg$ with 1 bit
\STATE $qc \gets$ QuantumCircuit$(qreg, creg)$

\STATE Apply RY gate on $qreg[0]$ with angle $2a$
\STATE Apply RY gate on $qreg[0]$ with angle $2b$

\STATE Add measurement of $qreg[0]$ to $creg$

\STATE $counts \gets$ simulate\_quantum\_circuit$(qc)$

\STATE $quantum\_gasf[i][j] \gets  counts$
\ENDFOR
\ENDFOR

\STATE Append $quantum\_gasf$ to $quantum\_gasfs$
\ENDFOR

\STATE {\bfseries End Algorithm}
\end{algorithmic}
\end{algorithm}

Our study introduces a methodology for training time series data utilizing a CNN, as delineated in Algorithm \ref{algorithm2}. The procedure commences with a given time series data X accompanied by its respective labels Y. With parameters such as window size w and sliding stride s, the data X is segmented into sequential portions of length w with a sliding window advancing by s points. For each of these segments, a QGASF image is constructed. Subsequently, a training dataset is formulated, pairing these QGASF images with their corresponding labels from Y. A CNN is then defined, and during each training iteration, a batch of this dataset is ingested. The training process encompasses forward propagation to obtain outputs, followed by loss computation, backpropagation, and parameter updating. The culmination of this methodology is a trained CNN model.

\begin{algorithm}[H]
\caption{CNN-Based Training of QGASF Images}
\label{algorithm2}
\begin{algorithmic}
\REQUIRE Time series data $X=\{x_1,x_2,...,x_n\}$, labels $Y=\{y_1,y_2,...,y_n\}$
\ENSURE CNN model
\STATE Parameters: Window size $w$, sliding stride $s$
\STATE Split $X$ into sequential segments, each of length $w$, with a sliding window of $s$ points.
\FOR{each segment in $segments$}
\STATE Construct a QGASF image $Q = \text{QGASF}(segment)$
\ENDFOR
\STATE Build a training dataset $\{(Q_1,y_1), (Q_2,y_2), ..., (Q_m,y_m)\}$ using labels $Y$.
\STATE Define a Convolutional Neural Network (CNN).
\FOR{each training iteration}
\STATE Read a batch of training data.
\STATE Perform forward propagation to obtain outputs.
\STATE Calculate the loss function, perform backpropagation, and update parameters.
\ENDFOR
\STATE Return the CNN model.
\end{algorithmic}
\end{algorithm}

\section{Classical GASF generating workflow\label{Appendix B}}

\begin{figure}[H]
\centering
\includegraphics[width=15.5cm]{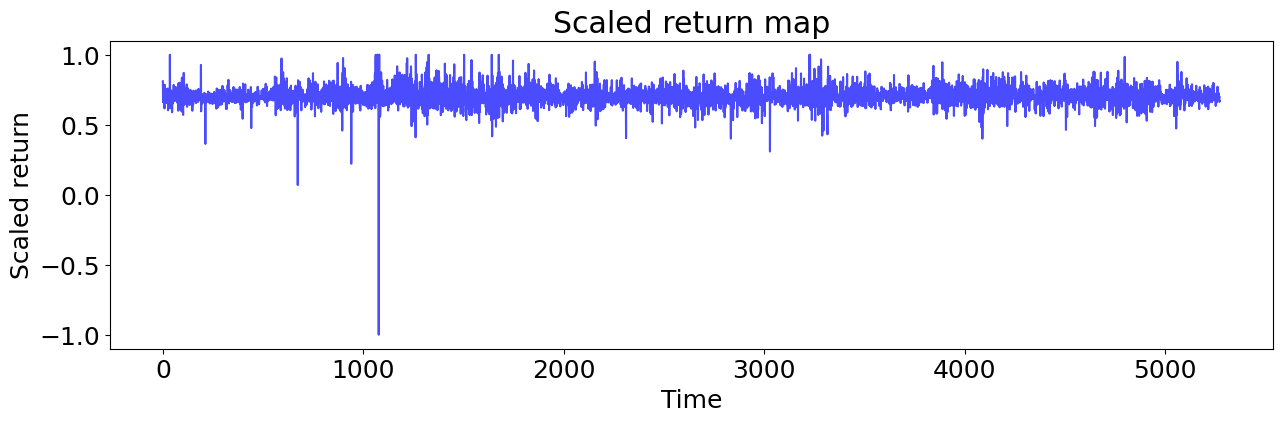}
\caption{ The effect of scaling the stock return data to between -1 and 1 according to equation \ref{eq2} in section \ref{sub3.2.1}.\label{figB.1}}
\end{figure}

\begin{figure}[H]
\centering
\includegraphics[width=10.5 cm]{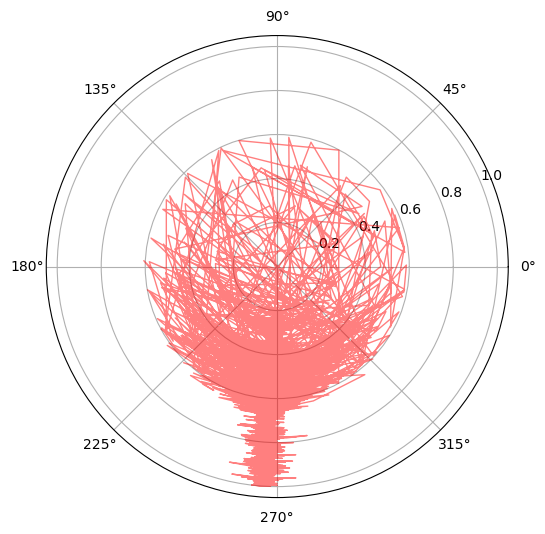}
\caption{A schematic representation of transforming raw time series data into polar coordinates, where the values of the time series and their corresponding timestamps are represented as angles and radii, respectively. Specific reference can be made to equations \ref{eq3} and \ref{eq4} in section \ref{sub3.2.1}.\label{figB.2}}
\end{figure}

\begin{figure}[H]
\centering
\includegraphics[width=15.5cm]{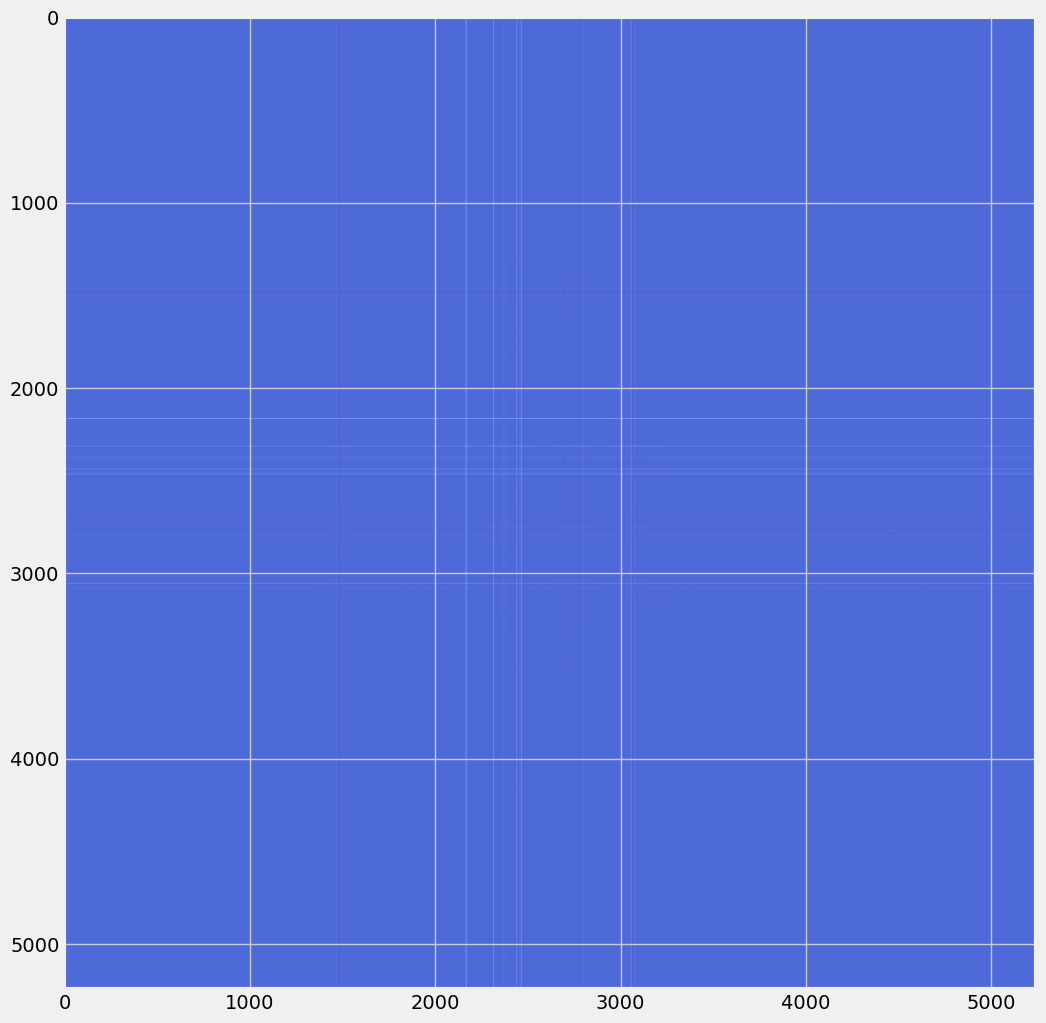}
\caption{All the values undergoing the inverse cosine transform are computed and iterated by cosine summation via equation \ref{eq5} in Section \ref{sub3.2.1} to all the pixel points of a new map containing information about the daily returns of the Moutai stock for more than 5000 days.\label{figB.3}}
\end{figure}

As described in Section \ref{sub3.2.1}, when generating classical GASF images, we first need to normalize the raw data using equation \ref{eq1} or \ref{eq2}, obtaining the scaled images as shown in the figure \ref{figB.1}. In addition, we can also use the equation \ref{eq3} and equation \ref{eq4} to convert the normalized time series data into a polar coordinate system (figure \ref{figB.2}) and ultimately use the equation \ref{eq5} to calculate the classical GASF image(figure \ref{figB.3}). Here, we chose Moutai stock as an example consistent with the main text.

\unskip

\section{MAE and MSE values}

The Tables \ref{tab1}-\ref{tab3} are the MAE and MSE values for three different stock markets corresponding to Figure \ref{fig9}-\ref{fig11} in the main text.

\begin{table}[h]
\caption{Model performance for the China A-share market.} \label{tab1}
\centering
\begin{tabular}{ccccccccccc}
\toprule
& \multicolumn{2}{c}{600030.SH} & \multicolumn{2}{c}{600036.SH} & \multicolumn{2}{c}{600519.SH} & \multicolumn{2}{c}{601166.SH} & \multicolumn{2}{c}{601318.SH} \\
\cmidrule(lr){2-3} \cmidrule(lr){4-5} \cmidrule(lr){6-7} \cmidrule(lr){8-9} \cmidrule(lr){10-11}
& \textbf{MAE} & \textbf{MSE} & \textbf{MAE} & \textbf{MSE} & \textbf{MAE} & \textbf{MSE} & \textbf{MAE} & \textbf{MSE} & \textbf{MAE} & \textbf{MSE} \\
\midrule
GASF & 0.18 & 0.04 & 0.105 & 0.018 & 0.195 & 0.032 & 0.115 & 0.028 & 0.115 & 0.026 \\
QGASF & 0.11 & 0.03 & 0.081 & 0.010 & 0.119 & 0.015 & 0.086 & 0.015 & 0.093 & 0.013 \\
\bottomrule
\end{tabular}
\end{table}

\begin{table}[h]
\caption{Model performance for the Hong Kong stock market.}
\label{tab2}
\centering
\begin{tabular}{lccccccccccc}
\toprule
& \multicolumn{2}{c}{0005.HK} & \multicolumn{2}{c}{0700.HK} & \multicolumn{2}{c}{0939.HK} & \multicolumn{2}{c}{0941.HK} & \multicolumn{2}{c}{1299.HK} \\
\cmidrule(lr){2-3} \cmidrule(lr){4-5} \cmidrule(lr){6-7} \cmidrule(lr){8-9} \cmidrule(lr){10-11}
& \textbf{MAE} & \textbf{MSE} & \textbf{MAE} & \textbf{MSE} & \textbf{MAE} & \textbf{MSE} & \textbf{MAE} & \textbf{MSE} & \textbf{MAE} & \textbf{MSE} \\
\midrule
GASF & 0.086 & 0.093 & 0.115 & 0.075 & 0.113 & 0.016 & 0.083 & 0.012 & 0.066 & 0.014 \\
QGASF & 0.061 & 0.010 & 0.092 & 0.017 & 0.066 & 0.009 & 0.069 & 0.010 & 0.046 & 0.005 \\
\bottomrule
\end{tabular}
\end{table}

\begin{table}
\caption{Model performance for the US stock market. \label{tab3}}
\centering
\begin{tabular}{lccccccccccc}
\toprule
& \multicolumn{2}{c}{BA} & \multicolumn{2}{c}{DOW} & \multicolumn{2}{c}{IBM} & \multicolumn{2}{c}{JPM} & \multicolumn{2}{c}{PG} \\
\cmidrule(lr){2-3} \cmidrule(lr){4-5} \cmidrule(lr){6-7} \cmidrule(lr){8-9} \cmidrule(lr){10-11}
& \textbf{MAE} & \textbf{MSE} & \textbf{MAE} & \textbf{MSE} & \textbf{MAE} & \textbf{MSE} & \textbf{MAE} & \textbf{MSE} & \textbf{MAE} & \textbf{MSE} \\
\midrule
GASF & 0.093 & 0.018 & 0.084 & 0.025 & 0.120 & 0.009 & 0.081 & 0.012 & 0.052 & 0.013 \\
QGASF & 0.090 & 0.014 & 0.079 & 0.011 & 0.062 & 0.005 & 0.077 & 0.009 & 0.037 & 0.004 \\
\bottomrule
\end{tabular}
\end{table}

\end{document}